\documentclass[letterpaper, 10 pt, conference]{ieeeconf}  

\IEEEoverridecommandlockouts                              

\usepackage[dvipsnames]{xcolor}

\usepackage{amsmath,amsfonts,amsthm,amssymb}
\usepackage{mathtools}
\usepackage{bm}
\usepackage{nicefrac}
\usepackage{microtype}

\usepackage{xcolor}
\usepackage{epsfig}
\usepackage{graphicx}

\usepackage{adjustbox}
\usepackage{array}
\usepackage{booktabs}
\usepackage{colortbl}
\usepackage{wrapfig}
\usepackage{hhline}
\usepackage{multirow}
\usepackage{subcaption}
\usepackage[size=small]{caption}

\usepackage{changepage}
\usepackage{extramarks}
\usepackage{fancyhdr}
\usepackage{lastpage}
\usepackage{setspace}
\usepackage{soul}
\usepackage{xspace}
\usepackage{adjustbox}

\usepackage{enumerate}
\usepackage{titlesec}
\usepackage{makecell}
\usepackage{pifont}

\usepackage{hyperref}
\usepackage{url}
\usepackage[capitalise]{cleveref}
\usepackage[linesnumbered,ruled,vlined]{algorithm2e}

\newcolumntype{L}[1]{>{\raggedright\let\newline\\\arraybackslash\hspace{0pt}}m{#1}}
\newcolumntype{C}[1]{>{\centering\let\newline\\\arraybackslash\hspace{0pt}}m{#1}}
\newcolumntype{R}[1]{>{\raggedleft\let\newline\\\arraybackslash\hspace{0pt}}m{#1}}

\newcommand{\ignore}[1]{}

\DeclareMathAlphabet{\mathbfit}{OML}{cmm}{b}{it}

\makeatletter
\DeclareRobustCommand\onedot{\futurelet\@let@token\@onedot}
\def\@onedot{\ifx\@let@token.\else.\null\fi\xspace}

\def\eg{\textit{e.g}\onedot} 
\def\ie{\textit{i.e}\onedot}

\makeatother

\definecolor{MyDarkBlue}{rgb}{0,0.08,1}
\definecolor{MyAqua}{rgb}{0,0.7,0.7}
\definecolor{MyDarkGreen}{rgb}{0.02,0.6,0.02}
\definecolor{MyDarkOrange}{rgb}{0.40,0.2,0.02}
\definecolor{MyPurple}{RGB}{111,0,255}
\definecolor{MyGold}{rgb}{0.75,0.6,0.12}
\definecolor{MyDarkgray}{rgb}{0.66, 0.66, 0.66}

\definecolor{MyRed}{rgb}{0.8,0.25,0.25}
\definecolor{MyGreen}{rgb}{0.25,0.8,0.25}
\definecolor{MyBlue}{rgb}{0.25,0.25,0.8}

\newcommand{\myparagraph}[1]{\vspace{0.1mm}\noindent\textbf{#1}}

\titlespacing*{\section}{2pt}{2pt plus 2pt minus 2pt}{1pt plus 2pt minus 2pt}
\titlespacing*{\subsection}{0pt}{0pt plus 1pt minus 1pt}{0pt plus 1pt minus 1pt}

\crefname{algorithm}{Alg.}{Algs.}
\Crefname{algocf}{Algorithm}{Algorithms}
\crefname{section}{Sec.}{Secs.}
\Crefname{section}{Section}{Sections}
\crefname{table}{Tab.}{Tabs.}
\Crefname{table}{Table}{Tables}
\crefname{figure}{Fig.}{Fig.}
\Crefname{figure}{Figure}{Figure}

\usepackage{amsmath,amsfonts,bm}

\def\eqref#1{equation~\ref{#1}}

\def\1{\bm{1}}

\DeclareMathAlphabet{\mathsfit}{\encodingdefault}{\sfdefault}{m}{sl}
\SetMathAlphabet{\mathsfit}{bold}{\encodingdefault}{\sfdefault}{bx}{n}

\def\ie{\textit{i.e.,~}}
\def\eg{\textit{e.g.,~}}

\usepackage{cleveref}
\usepackage{mwe}
\usepackage{xspace}
\usepackage{balance}
\usepackage{acronym}
\usepackage{ulem}
\usepackage{progressbar}
\usepackage{ifthen}
\usepackage{xparse}
\usepackage{xfrac}

\title{PhysPart: Physically Plausible Part Completion \\for Interactable Objects}

\author{
Rundong Luo$^{*1}$, Haoran Geng$^{*2,3}$, Congyue Deng$^{3}$, Puhao Li$^{4}$, \\Zan Wang$^{4}$, Baoxiong Jia$^{4}$, Leonidas Guibas$^{3}$, and Siyuan Huang$^{4}$%
\thanks{$^{1}$Cornell University,}%
\thanks{$^{2}$University of California, Berkeley}%
\thanks{$^{3}$Stanford University}%
\thanks{$^{4}$Beijing Institute for General Artificial Intelligence (BIGAI)}%
\thanks{$*$Equal Contribution}
}

\let\oldtwocolumn\twocolumn
\renewcommand\twocolumn[1][]{%
    \oldtwocolumn[{#1}{
        \centering
        \vspace{-5mm}
        \includegraphics[width=0.86\linewidth]{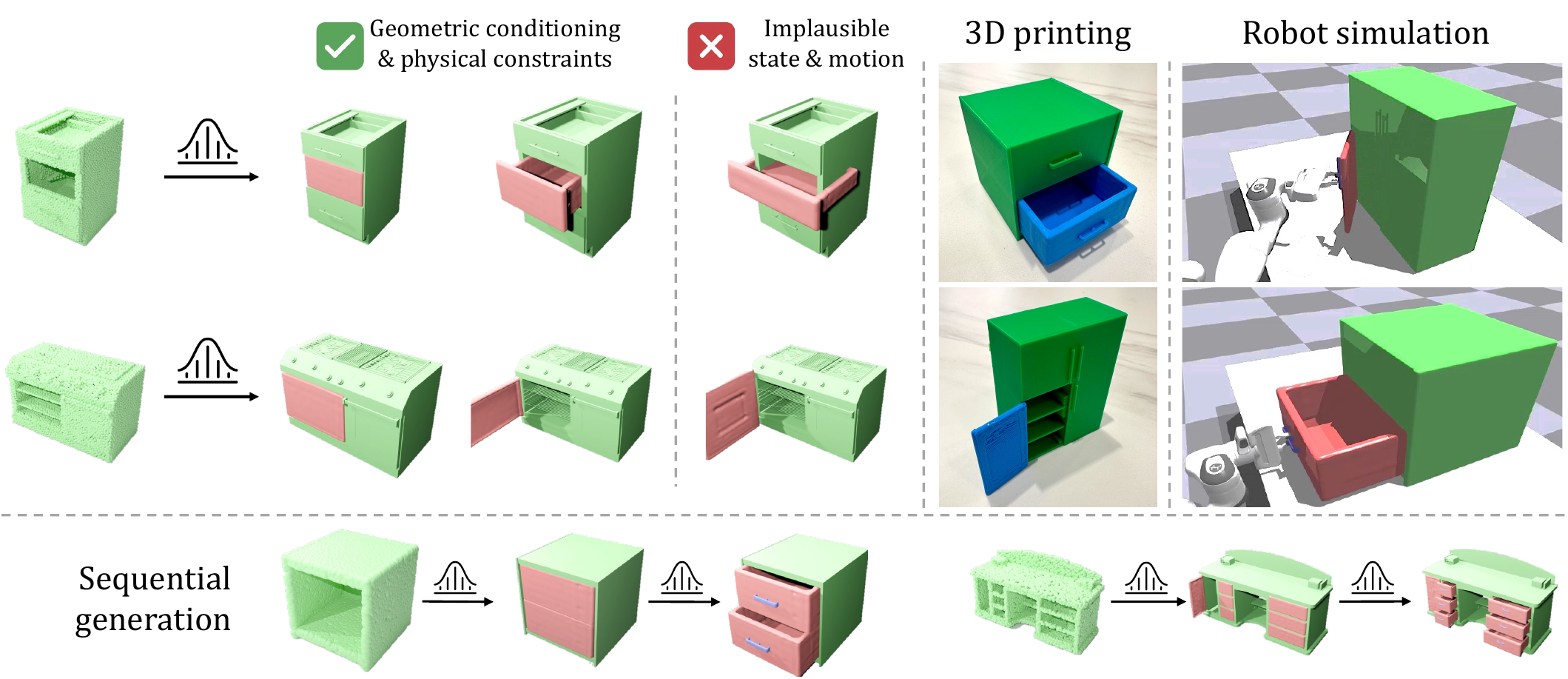}
        \vspace{-1mm}
        \captionof{figure}{\textbf{Left:} Given an incomplete object base, we propose a diffusion-based model to generate physically plausible parts that allow articulated interactions.
            \textbf{Right:} Our model naturally applies to real-world scenarios and embodied AI tasks, including 3D printing and robot manipulation, where physical plausibility is essential to the object interactivity and success of the tasks. For example, a small surface bump or a slight size mismatch can get the part stuck during interactions.
            \textbf{Bottom:} Our model supports sequential generation of dependent parts (\eg cabinet$\to$door$\to$handle) for more complex object hierarchies.}
        \label{fig:teaser}
    }]
}

\begin{document}

\maketitle
\thispagestyle{empty}
\pagestyle{empty}

\begin{abstract}
Interactable objects are ubiquitous in our daily lives.
Recent advances in 3D generative models make it possible to automate the modeling of these objects, benefiting a range of applications from 3D printing to the creation of robot simulation environments.
However, while significant progress has been made in modeling 3D shapes and appearances, modeling object physics, particularly for interactable objects, remains challenging due to the physical constraints imposed by inter-part motions.
In this paper, we tackle the problem of physically plausible part completion for interactable objects, aiming to generate 3D parts that not only fit precisely into the object but also allow smooth part motions. 
To this end, we propose a diffusion-based part generation model that utilizes geometric conditioning through classifier-free guidance and formulates physical constraints as a set of stability and mobility losses to guide the sampling process.
Additionally, we demonstrate the generation of dependent parts, paving the way toward sequential part generation for objects with complex part-whole hierarchies.
Experimentally, we introduce a new metric for measuring physical plausibility based on motion success rates. Our model outperforms existing baselines over shape and physical metrics, especially those that do not adequately model physical constraints.
We also demonstrate our applications in 3D printing, robot manipulation, and sequential part generation, showing our strength in realistic tasks with the demand for high physical plausibility.

\end{abstract}

\section{Introduction}

The creation of 3D objects that facilitate human interactions is a critical and long-standing problem in industrial engineering, with applications ranging from manufacturing to the development of simulation environments.
With the recent proliferation of AI-facilitated 3D content creation~\cite{poole2022dreamfusion, lin2023magic3d, deng2023nerdi, xu2023neurallift, zhou2023sparsefusion, liu2023zero, shi2023zero123++, liu2024one}, a line of efforts has been made to model interactable objects, particularly articulated ones, with 3D generative models~\cite{lei2023nap, liu2024cage}. Imagine having a cabinet with a broken drawer; rather than engaging in labor-intensive carpentry with jigs and saws, a more efficient solution would be to generate a 3D model of the drawer and print it using a 3D printer, thus eliminating the need for specialized human expertise.

However, unlike rigid objects, whose modeling primarily focuses on shapes and appearances, interactable object modeling presents the additional challenge of intricate part motions under human interactions, constrained by geometry and physics. This complexity is particularly problematic for real-world applications, where high accuracy and robustness are essential. For instance, in the case of a drawer, even a slight size mismatch or a minor surface bump can cause it to get stuck in the cabinet.

In this paper, we address the problem of generating 3D object parts that are physically plausible both in their rest states and under interactions. Previous work on physics-constrained shape generation has predominantly concentrated on rigid objects, emphasizing connectivity~\cite{mezghanni2021physically, hu2024topology, huang2024enhancing} or structural stability~\cite{mezghanni2022physical}. However, for interactable objects with part motions, a significant challenge arises: while the generated shapes are static, the inherent state changes are crucial for physical plausibility. Consequently, both ``first-order'' constraints on static states and ``second-order'' constraints on state changes must be carefully considered.

We propose a diffusion-based framework for physically plausible part completion for interactable objects.
Given an object point cloud with a missing part, we represent the part as an implicit surface and generate its shape with a latent diffusion model on quantized latent embeddings,
To guide the diffusion sampling process toward a physically plausible solution, we introduce two types of conditions: the \emph{geometric conditioning} applied via classifier-free guidance and the \emph{physical constraints} formulated as inference-time losses applied to the score functions.
Specifically, we design two sets of losses for physical constraints: stability losses that consider collisions and contacts at rest states and mobility losses that consider part motions.
To incorporate the physical losses into the sampling process, we approximate their gradients using a prediction of the noise-free sample at each time step and add the approximated gradients to the score function.
Finally, we also demonstrate the generation of dependent parts, paving the way toward sequential part generation for objects with complex part-whole hierarchies.

For evaluation, we construct a benchmark with metrics for assessing both the shapes and the physics of the generation results. Besides the widely adopted metrics for 3D generative models, we introduce a \emph{physical plausibility metric} that directly evaluates the part motions.
With both qualitative and quantitative results, we show that our framework generates high-quality shapes, outperforming methods that lack proper modeling of geometric conditioning or physical constraints.
Furthermore, we demonstrate 3D printing and robot manipulation applications, showcasing our strong potential in real-world applications that require high physical plausibility.

To summarize, our key contributions are:
\begin{itemize}
    \item We introduce the task of physically plausible part completion for interactable objects, along with a benchmark that includes comprehensive evaluation metrics for both shape and physical properties.
    \item We propose a diffusion-based part generative model that incorporates geometric conditioning and physical constraints, formulated as stability and mobility losses. Our model achieves superior results in both shape quality and physical plausibility compared to baseline models, demonstrating its effectiveness.
    \item We highlight physics-intensive applications of our model, such as 3D printing, robotic manipulation, and sequential part generation, where high physical realism is essential.
\end{itemize}
\section{Related Work}

\myparagraph{Articulated Object Modeling.}
Articulated object modeling is a crucial and longstanding field in 3D vision and robotics, encompassing a wide range of work in perception~\cite{geng2022gapartnet, yi2018deep, deng2024banana, li2020category, liu2023semi, geng2024sage, geng2022end, geng2023partmanip, gong2023arnold}, reconstruction~\cite{chen2023urdformer, mu2021sdf, jiang2022ditto, tseng2022cla}, and generation~\cite{lei2023nap, liu2024cage}. These contributions have greatly advanced applications in simulation~\cite{gong2023arnold, yang2024physcene} and robot manipulation~\cite{xu2022universal, cheng2023learning, ding2024opendor, yu2024dexgraspnet, li2024ag2maniplearningnovelmanipulation, geng2023partmanip, geng2024sage}. 
Beyond these perspectives, the intricate per-part state changes and inter-part motions, which formulates how objects interact and respond to interactions, also play important roles. For instance, \cite{geng2022gapartnet} categorizes parts based on their actions, \cite{yi2018deep} learns part segmentation by analyzing point clouds during part motions, and \cite{mu2021sdf, jiang2022ditto, tseng2022cla} focus on reconstructing parts and joints across various motion states.
Most studies focus on part motions at the pose level, often overlooking the crucial interaction between physics and detailed part geometry. This gap is particularly important in 3D generation, where physical plausibility is key for practical use. However, recent research has under-explored this aspect, concentrating more on structure and shape~\cite{lei2023nap, liu2024cage}. For a more comprehensive overview, refer to \cite{liu2024survey}.

\myparagraph{Physics-Aware Shape Generation.}
Previous research in physics-aware 3D object generation has primarily focused on rigid shapes without complex self-interactions.
\cite{mezghanni2021physically, hu2024topology, huang2024enhancing} have used persistence diagrams to ensure topological connectivity, and \cite{mezghanni2022physical} introduced a differentiable physical simulation layer to enhance stability under gravity. In contrast, physical principles have been extensively applied in other areas, such as in PhyScene~\cite{yang2024physcene}, which aids in synthesizing interactable 3D scenes, and in human pose and motion generation, where studies~\cite{zhang2020place, zhang2020generating, wang2021synthesizing} incorporate physics-aware constraints to reduce collisions and enhance realism. Additionally, built on diffusion models, SceneDiffuser~\cite{huang2023diffusion} and PhysDiff~\cite{yuan2023physdiff} achieve physically plausible generation by incorporating physical optimization into the denoising process. Our approach extends this by embedding physical constraints into a generative diffusion model, enabling the creation of physically plausible parts.

\begin{figure*}[!t]
    \centering
    \vspace{-3mm}
    \includegraphics[width=0.85\linewidth]{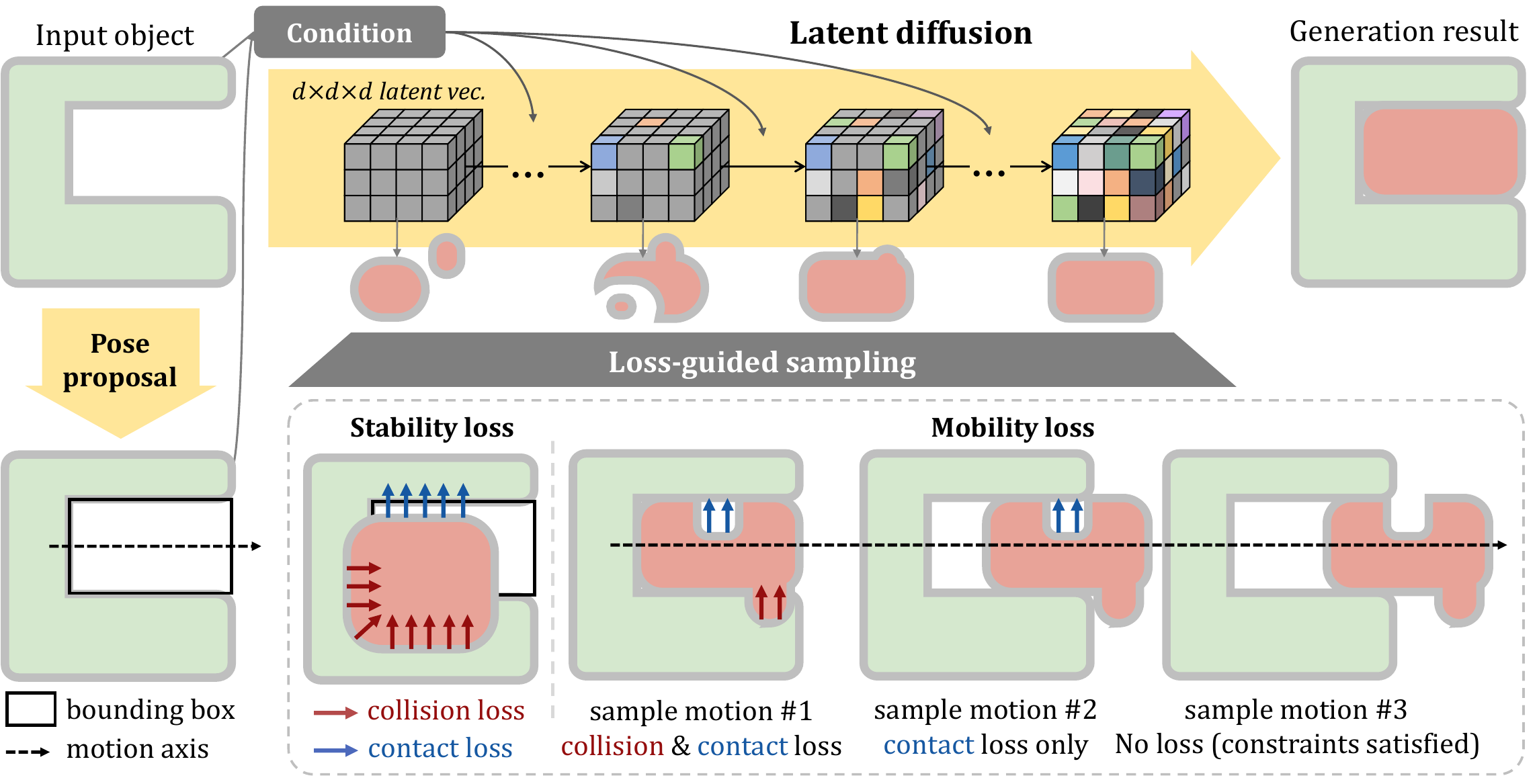}
    \vspace{-2mm}
    \caption{\textbf{Pipeline of our proposed framework}. We train a pose proposal model to predict the missing part's bounding box and a latent diffusion model conditioned on the input object's point cloud and the missing part's bounding box within the latent space of part SDF. During inference, the trained pose proposal model first predicts the missing part's bounding box. We then apply the proposed physical-aware losses (contact and collision losses in static or dynamics states) to guide the sampling process.}
    \vspace{-6mm}
    \label{fig:pipeline}
\end{figure*}

\myparagraph{3D Shape Completion.} Shape completion is a vital step for reconstructing the missing parts of 3D shapes, traditionally tackled with techniques such as Laplacian hole filling~\cite{sorkine2004least, nealen2006laplacian, zhao2007robust} and Poisson surface reconstruction~\cite{kazhdan2006poisson, kazhdan2013screened}. These methods primarily address small gaps and basic geometric forms. Another strategy employs structural regularities, like symmetries, to infer unobserved parts of shapes~\cite{thrun2005shape, mitra2006partial, pauly2008discovering, sipiran2014approximate, speciale2016symmetry}. With the advent of extensive 3D datasets, retrieval-based approaches~\cite{sung2015data, li2015database, nan2012search, kim2012acquiring} have emerged, searching databases to find the best matches for incomplete inputs, alongside learning-based methods~\cite{nguyen2016field, firman2016structured, dai2017shape, dai2020sg, yu2021pointr, han2017high, song2017semantic, chibane2020implicit} that minimize discrepancies between network predictions and actual shapes. Notably, 3D-EPN~\cite{dai2017shape} utilizes a 3D encoder-decoder architecture to predict complete shapes from partial volumetric data, while Scan2Mesh~\cite{dai2019scan2mesh} transforms range scans into 3D meshes through direct mesh surface optimization. PatchComplete~\cite{rao2022patchcomplete} leverages local structural priors to fill in shapes from unseen categories. Alternatively, generative methods like GANs~\cite{zheng2022sdf, zhang2021unsupervised, chen2019unpaired, wu2020multimodal, smith2017improved} and AutoEncoders~\cite{mittal2022autosdf, achlioptas2018learning} offer a different approach by generating diverse, plausible shapes from partial inputs, though they often compromise on completion accuracy. Our method leverages diffusion models and physics-aware loss guidance, effectively reducing surface artifacts and surpassing current state-of-the-art methods in producing realistic and physics-plausible 3D shapes.

\section{Method}

Given the full point cloud of an articulated object $\mathcal{O}$ that has a missing part, we aim to generate a replacement part $p$ that integrates seamlessly into the object while adhering to geometric and physical constraints. 
We employ volumetric Truncated Signed Distance Fields (T-SDF) to model object parts and train a 3D-VQVAE to compress these models into a discrete and compact latent space.
Unlike prior research that employs SDF representation that often overlooks the object's scale as generating SDF usually involves normalizing the part to the unit cube, our framework also emphasizes accurate scaling of the objects for manufacturing purposes.
Upon this part representation, our method utilizes a coarse-to-fine approach, involving two diffusion-based sub-models: a pose proposal model that determines the part’s approximate position at the coarse level, and a part code generator that refines the shape based on geometric conditions (\cref{sec:method:geometry}) and physical constraints~(\cref{sec:method:physics}). This integrated framework ensures that the generated part can interact properly with external forces and fit accurately within the existing object structure. An overview of our pipeline is shown in \cref{fig:pipeline}.

\myparagraph{Part-Pose Proposal.}
We propose a diffusion-based pose estimator to determine the bounding box of the part to be generated from the object point cloud $\mathcal{P}$. Drawing inspiration from GenPose~\cite{zhang2023genpose}, we train a score-based diffusion model $\Phi_\theta$ alongside an energy-based diffusion model $\Psi_\phi$. First, we generate pose candidates ${p^i}{i=1}^K$ using $\Phi\theta$. Then, the pose energies $\Psi_\phi(p^i, \mathcal{P})$ are computed for each candidate via $\Psi_\phi$. By ranking the candidates based on their energies and filtering out the lower-ranked ones, the remaining bounding box candidates are aggregated into the final output using mean pooling.

\begin{table*}[t]
    \scriptsize
    \centering
    \vspace{-4mm}
    \caption{Quantitative generation results on self-moving parts. ``Phys.'' denotes our percentage physical plausibility metric. ``Ours, w/o loss-guided samp.'' denotes our method without using the proposed loss-guided sampling. }
     \vspace{-0.5mm}
    \begin{tabular}{l  ccc  ccc  ccc  ccc}
    \toprule \vspace{0.2mm}
    & \multicolumn{6}{c}{\textbf{Self-moving parts}} & \multicolumn{6}{c}{\textbf{Dependent parts}} \\ \midrule 
    \multirow{2}{*}{Method} & \multicolumn{3}{c}{Slider drawer} & \multicolumn{3}{c}{Hinge door} &  \multicolumn{3}{c}{Line handle} & \multicolumn{3}{c}{Hinge knob} \\ \cmidrule(lr){2-4}\cmidrule(lr){5-7}\cmidrule(lr){8-10}\cmidrule(lr){11-13}
      & CD$\downarrow$ & F-score$\uparrow$ & Phys. & CD$\downarrow$ & F-score$\uparrow$  & Phys. & CD$\downarrow$ & F-score $\uparrow$ & Phys. & CD$\downarrow$ & F-score$\uparrow$ & Phys. \\ \midrule
      C-VAE  & 0.00248 & 0.683 & 38.9 & 0.00106 & 0.867 & 45.6 & 0.01000 & 0.403 & 63.3 & 0.00028 & 0.894 & 31.5 \\ \midrule
       Ours, w/o loss-guided samp. & 0.00082 & 0.818 & 49.3 & 0.00083 & 0.873 & 71.1 & 0.00174 & 0.742 & 71.1 & 0.00026 & 0.912 & 78.9 \\ \midrule
       \textbf{Ours} & \textbf{0.00060} & \textbf{0.862} & \textbf{74.3} & \textbf{0.00081} & \textbf{0.883} & \textbf{92.2} & \textbf{0.00173} & \textbf{0.744} & \textbf{85.9} & \textbf{0.00024} & \textbf{0.917} & \textbf{84.2}\\
      \bottomrule
    \end{tabular}
    \label{tab:quantative}
    \vspace{-5mm}
\end{table*}

\subsection{Geometry-Conditioned Part Generation}
\label{sec:method:geometry}

\paragraph{SDF Generation with Latent Diffusion.} SDFusion~\cite{cheng2023sdfusion} introduces a diffusion-based 3D generative model. The approach begins by compressing high-dimensional SDFs into a compact latent space using a 3D-VQVAE. Specifically, an SDF $X \in \mathbb{R}^{D \times D \times D}$ is encoded into a latent vector $z \in \mathbb{R}^{d \times d \times d}$ by the encoder $E_{\phi}$. The encoded SDF is then reconstructed by the decoder $D_{\tau}$ after passing $z$ through a vector quantization codebook VQ, \textit{i.e.,}
\begin{equation}
z = E_{\phi}(X), \text{ and } X' = D_{\tau}(\text{VQ}(z)).
\end{equation}
Afterwards, conditional diffusion model is trained on this learned latent distribution. The classifier-free guidance~\cite{ho2022classifier} and DDIM~\cite{song2020denoising} sampling techniques are employed to enhance the flexibility and efficiency of the generation process.

\myparagraph{Geometric Conditioning.} The above framework allows the generation of missing parts using point cloud as the condition. However, due to limited training data and the point cloud's inherent ambiguity, relying solely on the point cloud often yields unsatisfactory results. To improve the model's performance, we leverage the part bounding box predicted by our pose estimation model. Specifically, given the predicted bounding box of the desired part, we rotate to canonical space and employ the ratio of the dimensions along the three axes as an auxiliary condition for the diffusion model. We use the length ratio rather than absolute values since SDF representation is scale-agnostic. Note that although we allow rotation to the input point cloud, the mesh is always generated in its canonical pose.

Empirically, we adopt the PointNet++~\cite{qi2017pointnet++}~($\phi_1$) to encode the point cloud~$c_1\in\mathbb{R}^{N\times3}$, and a sinusoidal positional encoding followed by a two-layer MLP ($\phi_2$) to encode the bounding box ratio~$c_2\in\mathbb{R}^{3}$. The training objective for the diffusion model $\theta$ and encoders $\phi_1, \phi_2$ is formulated as:
\begin{align}
    L(\theta, &\phi_1, \phi_2) := \notag \\  &\mathbb{E}_{z,c,e,t} \left[ \left\| \epsilon - \epsilon_\theta(z_t, t, D(\phi_1(c_1)), D(\phi_2(c_2))) \right\|^2 \right],
\end{align}
where $D(\cdot)$ denotes a dropout operation, $c_1, c_2$ are the conditions, and $z_t$ is the noisy latent variable at timestamp $t$. 

Finally, using the predicted bounding box from the pose proposal model, we scale the mesh by the cube root of the volume ratio between the predicted bounding box and the axis-aligned bounding box of the generated mesh. This ensures the mesh is sized appropriately to fit the object's empty region.

\begin{figure}[!t]
    \centering
    \includegraphics[width=\linewidth]{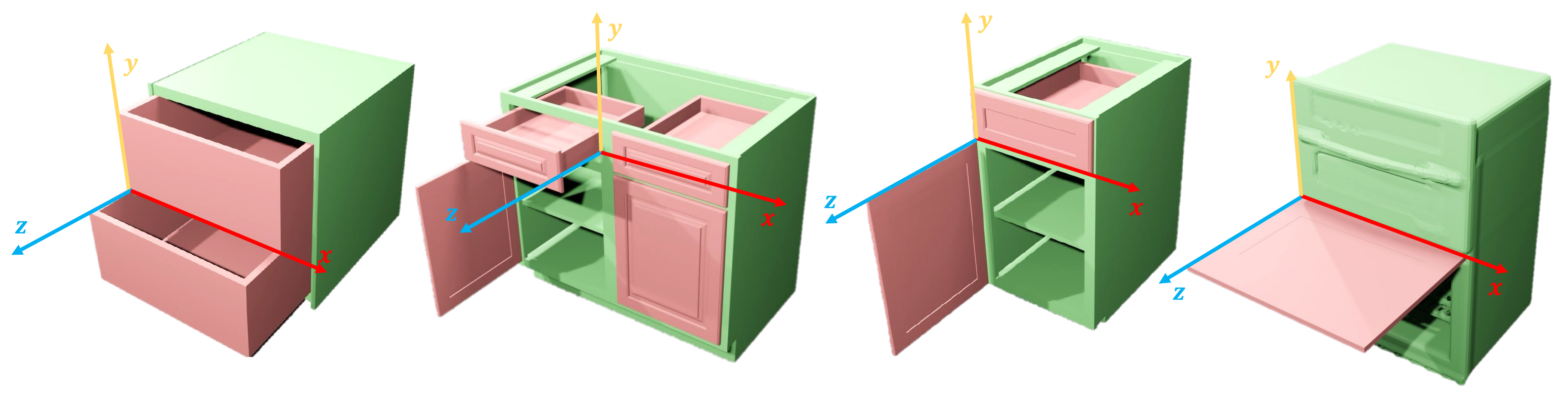}
    \vspace{-6mm}
    \caption{\textbf{Normalized 3D coordinate and shared motion constraints for certain part categories.} Specifically, drawers could be pulled out along the $+z$ axis, while doors could rotate around $+x$, $-x$, $+y$, and $-y$ axes.}
    \vspace{-7mm}
    \label{fig:3d-convention}
\end{figure}

\subsection{Loss-Guided Physical Constraints}
\label{sec:method:physics}

Given an object's point cloud, we predict the bounding box for the missing part, generate the desired part by sampling from the diffusion model using both the object's point cloud and the bounding box as conditions, and scale the generated mesh using the bounding box volume ratio. However, this generation process is agnostic to physical constraints and may result in parts incompatible with the original object.
To address this, we introduce physics-aware losses to guide the diffusion sampling process, ensuring physical compatibility.

\noindent\textbf{Stability Losses.}
The most critical criterion for the generated part is to avoid collision with the provided object point cloud $\mathcal{P}$. Therefore, we define the collision loss by considering the predicted part-SDF value $X(p)$ of the object point $p$:
\begin{equation}
\ell_{coll}(X, \mathcal{P}) = \sum_{p \in \mathcal{P}} \text{ReLU}(-X(p)-\alpha),
\label{l-col}
\end{equation}
where $\alpha$ represents a margin of tolerance. This loss function penalizes the SDF values when object points are inside the predicted SDF (\ie $X(p) < 0$), ensuring no parts of the object intersect the generated part. Additionally, maintaining proximity between the generated part and the object is crucial. Thus, we define contact loss to ensure no part of the object is beyond a distance $\alpha$ from the generated part:
\begin{equation}
\ell_{cont}(X, \mathcal{P}) = \min_{p \in \mathcal{P}} \text{ReLU}(X(p)-\alpha),
\label{l-cont}
\end{equation}
which penalizes excessive distances between the object points and the predicted SDF. Given that all parts are generated in a scale-agnostic canonical space, we adjust the scale of each part to match the predicted volume and apply transformations according to the predicted pose during loss calculations.

\begin{table*}[!t]
    \scriptsize
    \centering
    \vspace{-4mm}
    \caption{Ablation studies on the pose proposal model and conditional generation model.}
     \vspace{-1mm}
    \begin{tabular}{l  ccc  ccc  ccc  ccc}
    \toprule \vspace{0.2mm}
    & \multicolumn{6}{c}{\textbf{Self-moving parts}} & \multicolumn{6}{c}{\textbf{Dependent parts}} \\ \midrule 
    \multirow{2}{*}{Method} & \multicolumn{3}{c}{Slider drawer} & \multicolumn{3}{c}{Hinge door} &  \multicolumn{3}{c}{Line handle} & \multicolumn{3}{c}{Hinge knob} \\ \cmidrule(lr){2-4}\cmidrule(lr){5-7}\cmidrule(lr){8-10}\cmidrule(lr){11-13} 
      & CD$\downarrow$ & F-score$\uparrow$ & Phys. & CD$\downarrow$ & F-score$\uparrow$  & Phys. & CD$\downarrow$ & F-score $\uparrow$ & Phys. & CD$\downarrow$ & F-score$\uparrow$ & Phys. \\  \midrule
      Regress. pose proposal & \textbf{0.00059} &	0.847 & 73.6 & 0.00134 & 0.736 & 50.0 & 0.00252 & 0.655 & 74.2 & 0.00036 & 0.847 & 81.1 \\ \midrule
      w/o bbox condition & 0.00178 & 0.740 & 53.5 & 0.00433 & 0.791 & 74.4  & 0.00784 & 0.572 & 67.0 & 0.00033 & 0.885 & 80.7 \\ \midrule
      Full model & 0.00060 & \textbf{0.862} & \textbf{74.3} & \textbf{0.00081} & \textbf{0.883} & \textbf{92.2} & \textbf{0.00173} & \textbf{0.744} & \textbf{85.9} & \textbf{0.00024} & \textbf{0.917} & \textbf{84.2} \\
      \bottomrule
    \end{tabular}
    \label{tab:ablation}
    \vspace{-6mm}
\end{table*}

\noindent\textbf{Mobility Losses.}
As shown in Fig.~\ref{fig:3d-convention}, certain part categories must follow shared motion constraints. For instance, drawers can move ``out,'' and hinge doors can rotate around one of the four potential rotation axes. Thus, we extend both stability losses with motion. Since the joints of these parts have only one degree of freedom, we could ensure the feasibility of consistent part motion by random sampling the part states. The enhanced loss functions are defined as follows:
\begin{align}
&\ell_{coll-m}(X, \mathcal{P}) = \mathbb{E}_{\hat{X} \sim M(X)} \sum_{p \in \mathcal{P}}  \text{ReLU}(-\hat{X}(p)-\alpha), \label{eq:collision-loss}\\
&\ell_{cont-m}(X, \mathcal{P}) = \mathbb{E}_{\hat{X} \sim M(X)} \min_{p \in \mathcal{P}} \text{ReLU}(\hat{X}(p)-\alpha).
\label{eq:contact-loss}
\end{align}
where $M(X)$ is a set of transformed SDF. Note that for hinge doors, the ground truth rotation axis is unavailable during inference. However, as all four potential axes impose similar constraints when rotating from 0 to 90 degrees, we opt for a shared axis for all hinge door instances. To ensure both losses are differentiable, we opt to transform the point cloud instead of transforming the SDF directly.

\noindent\textbf{Loss-Guided Sampling.} Conditional diffusion models aim to generate samples $x_0$ given the condition $y$, where the conditional score at timestamp $t$ is derived using Bayes' rule:
\begin{align}
    &\nabla_{x_t}\log{p_t}(x_t|y) = \nabla_{x_t} \log{p_t}(x_t) + \nabla_{x_t} \log{p_t}(y|x_t) \notag \\ 
    & = -\frac{1}{\sqrt{1 - \overline{\alpha}_t}}\epsilon_\theta(x_t, t) + \nabla_{x_t} \log{p_t}(y|x_t) \notag  \\ 
    & = -\frac{1}{\sqrt{1 - \overline{\alpha}_t}} (\epsilon_\theta(x_t, t) - \sqrt{1 - \overline{\alpha}_t}\cdot \nabla_{x_t} \log{p_t}(y|x_t)). 
\end{align}

Hence, the $\omega$-weighted modified score $\overline{\epsilon}_\theta$ is expressed as:
\begin{align}
   \overline{\epsilon}_\theta(x_t, t, y) =  \epsilon_\theta(x_t, t) - \sqrt{1 - \overline{\alpha}_t}\omega\cdot\nabla_{x_t} \log{p_t}(y|x_t),
\end{align}

where the gradient term $\nabla_{x_t} \log{p_t}(y|x_t)$ is written as $-\frac{1}{\sqrt{1 - \overline{\alpha}_t}} (\epsilon_\theta(x_t, t, y) - \epsilon_\theta(x_t, t))$ in classifier-free guidance.

To introduce physical constraints into the sampling process, we approximate the gradient using our defined physics-aware loss functions. As loss functions are defined on noise-free samples, we compute the loss on the predicted noise-free sample at time step $t$, denoted as $\hat{x}_0$:
\begin{align}
    \nabla_{x_t}\log{p_t}(y|x_t) &\approx \nabla_{x_t} \log{p_t}(y|\hat{x}_0) \notag \\ &=\nabla_{x_t} \log \frac{\exp (-\ell_y(\hat{x}_0))}{Z} = -\ell_y(\hat{x}_0),
\end{align}
where $Z$ is the normalizing constant independent of $x_t$.

Thus, we combine classifier-free guidance and loss guidance to obtain the modified score:
\begin{equation}
    \overline{\epsilon}_\theta(x_t, t, y) = \epsilon_\theta(x_t, t) + \omega_1 (\epsilon_\theta(x_t, t, y) - \epsilon_\theta(x_t, t)) + \omega_2 \ell_y(\hat{x}_0),
\end{equation}
where the $\omega_1, \omega_2$ are guidance weights. Denote $X_0$ to be the decoded SDF of latent code $x_0$, the $\ell_y(x_0)$ is the sum of our defined physical-aware losses:
\begin{equation}
    \ell_y(x_0) = 
    \ell_{coll} + \ell_{cont} + \ell_{coll-m} + \ell_{cont-m}.
\end{equation}
\vspace{-2mm}
\setlength{\textfloatsep}{0pt}
\begin{algorithm}[!t]
\vspace{-0.5mm}
\footnotesize
\SetKwInOut{Input}{input}\SetKwInOut{Output}{output}\SetKw{Return}{return}\SetKwInOut{Initiate}{initiate}\SetKw{False}{false}\SetKw{True}{true}\SetKw{And}{and}\SetKw{Exists}{exists}\SetKw{Where}{where}\SetKw{Or}{or}
    \Input{Object point cloud $\mathcal{P}$, mesh of the predicted missing part $x$, set of position candidate $C_1$ determined by grid search, set of small movement $C_2$, and loss margin $\beta$.}
    \Output{Whether mesh $x$ is physically plausible}
    \Initiate{$C_0=\emptyset$}
    \BlankLine
    Remove components in $x$ with volume $< 10^{-7}$ \;
    \If{$x$ \text{contains multiple components or is not watertight}}{
    \Return \False\;
    }
    Transform the mesh $x$ into SDF representation $X$\;
     \ForAll{position candidate $c \in C_1$}{
        $P^c \leftarrow P-c$;~~\tcp{translate point cloud by $-c$}
        \If{$\ell_{collision-m}(X,P^{c})<\beta$ \And $\ell_{contact-m}(X,P^{c})<\beta$
        }{Add $c$ to $C_0$\;}
    }
    \ForAll{$c' \in C_2$}{\If{\Exists $c \in C_0$ \Where ($\ell_{collision-m}(X,P^{c+c'})\geq\beta$ \Or $\ell_{contact-m}(X,P^{c+c'})\geq\beta$)}{\Return \True\;}}
    \Return \False\;
    \caption{Physical Plausibility Metric}
    \label{algo:physical-plausible}
    \vspace{-1mm}
\end{algorithm}

\section{Experiments} \label{sec:experiments}

This section demonstrates the effectiveness of our method in generating four types of common articulated objects: slider drawers, hinge doors, hinge knobs, and line handles, where the first two allow \textit{self-movements} and the latter two must be attached to other parts (denoted as \textit{dependent} parts). We then explore its potential applications in 3D printing, robot manipulation, and sequential part generation tasks. 

\myparagraph{Data and Implementation.}  We train our models on the GAPartNet dataset~\cite{geng2022gapartnet}, a part-centric object model dataset with part-level annotations and actionability alignment across part classes. Approximately 90\% of instances are allocated for training for each part category, with the remainder reserved for evaluation. All part categories utilize the same VQVAE model trained across all part categories at an input SDF resolution of 128, while a separate diffusion model is trained for each category. Both our method and all baselines involve random rotation of input point clouds around the vertical axis during training for fair comparisons. For testing, we sample three different rotation angles for each object instance. 

\begin{figure}[!t]
    \centering
    \includegraphics[width=\linewidth]{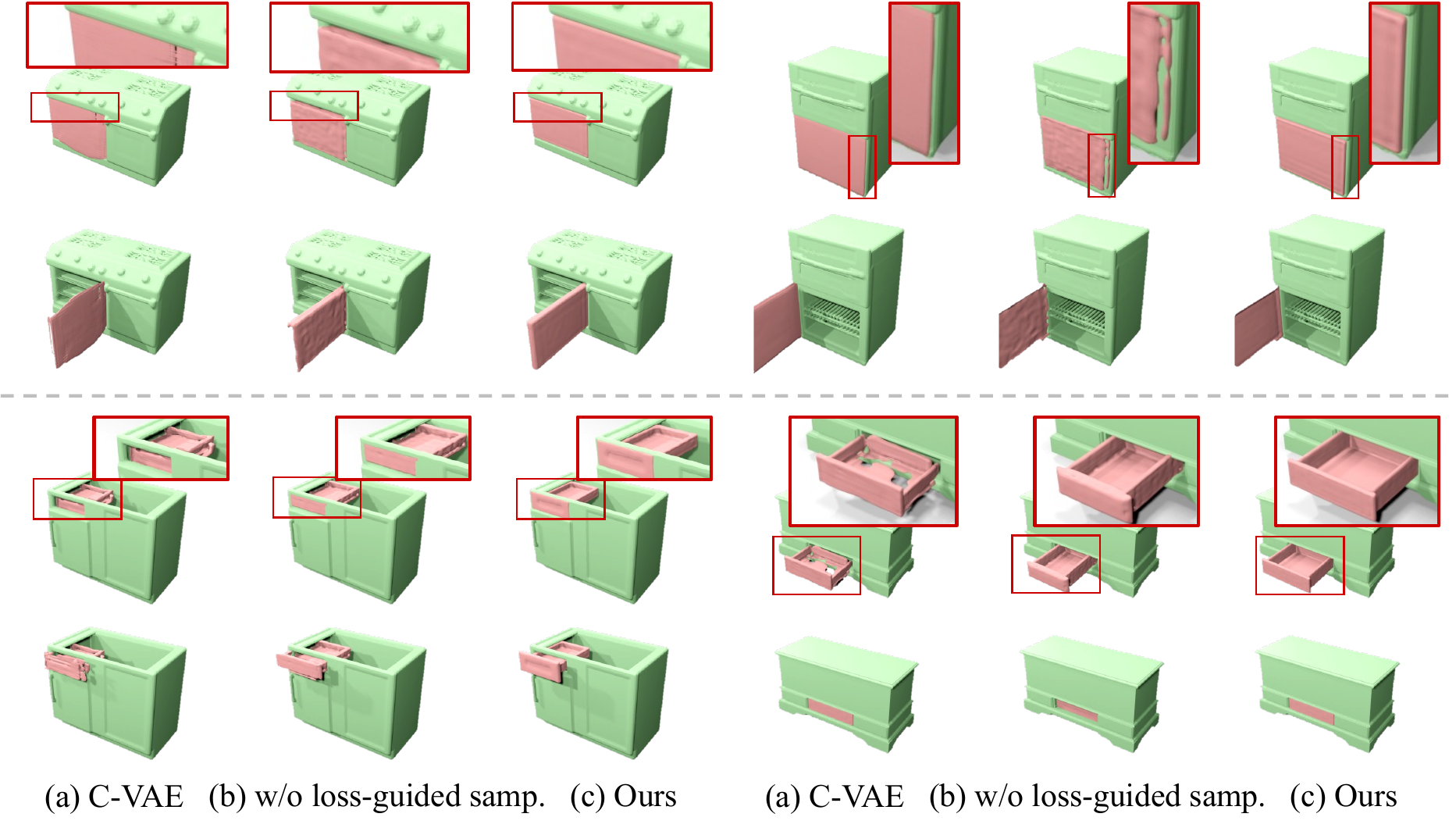}
    \vspace{-5mm}
    \caption{
    \textbf{Results of self-moving part generation.}
    All visualizations are done in the objects' canonical poses, with the base shapes in {\color{MyGreen}green} and the generated parts in {\color{MyRed}red}.
    While all methods can generate part shapes that are \emph{roughly} reasonable, the bumpy surfaces and size mismatches in baseline results hinder their \emph{physical plausibility}.
    }
    \vspace{-3mm}
    \label{fig:drawer-door}
\end{figure}

\begin{figure}[!t]
    \centering
    \includegraphics[width=\linewidth]{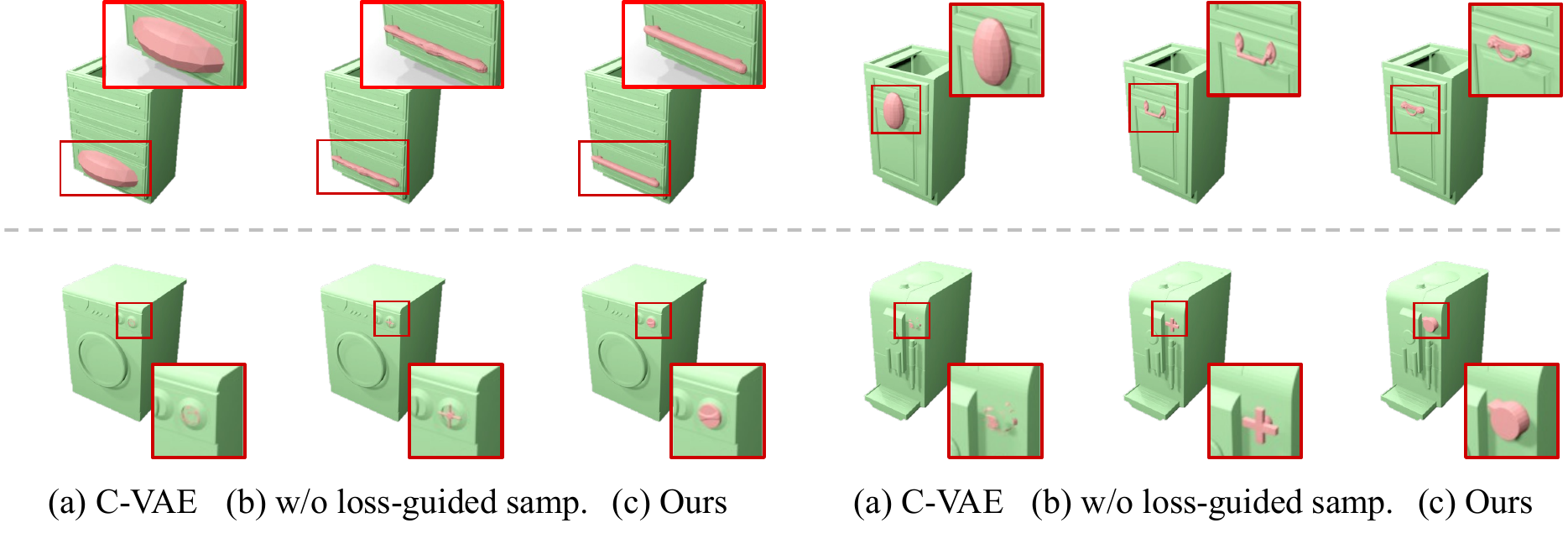}
    \vspace{-5mm}
    \caption{
    \textbf{Results of dependent part generation.}
    All visualizations are done in the objects' canonical poses, with the base shapes in {\color{MyGreen}green} and the generated parts in {\color{MyRed}red}.
    }
    \vspace{-1mm}
    \label{fig:handle-knob}
\end{figure}

\myparagraph{Baselines.} Our primary baseline is the Conditional Variational Autoencoder (C-VAE)~\cite{cvae}, which similarly utilizes a scale-agnostic SDF representation. To ensure fair comparisons, we scale the generated parts using our predicted pose and then use the same settings to process the results. We also include our method's performance without the proposed loss-guided sampling to demonstrate its effectiveness.

\begin{figure*}[!t]
    \centering
    \vspace{-5mm}
    \includegraphics[width=0.85\linewidth]{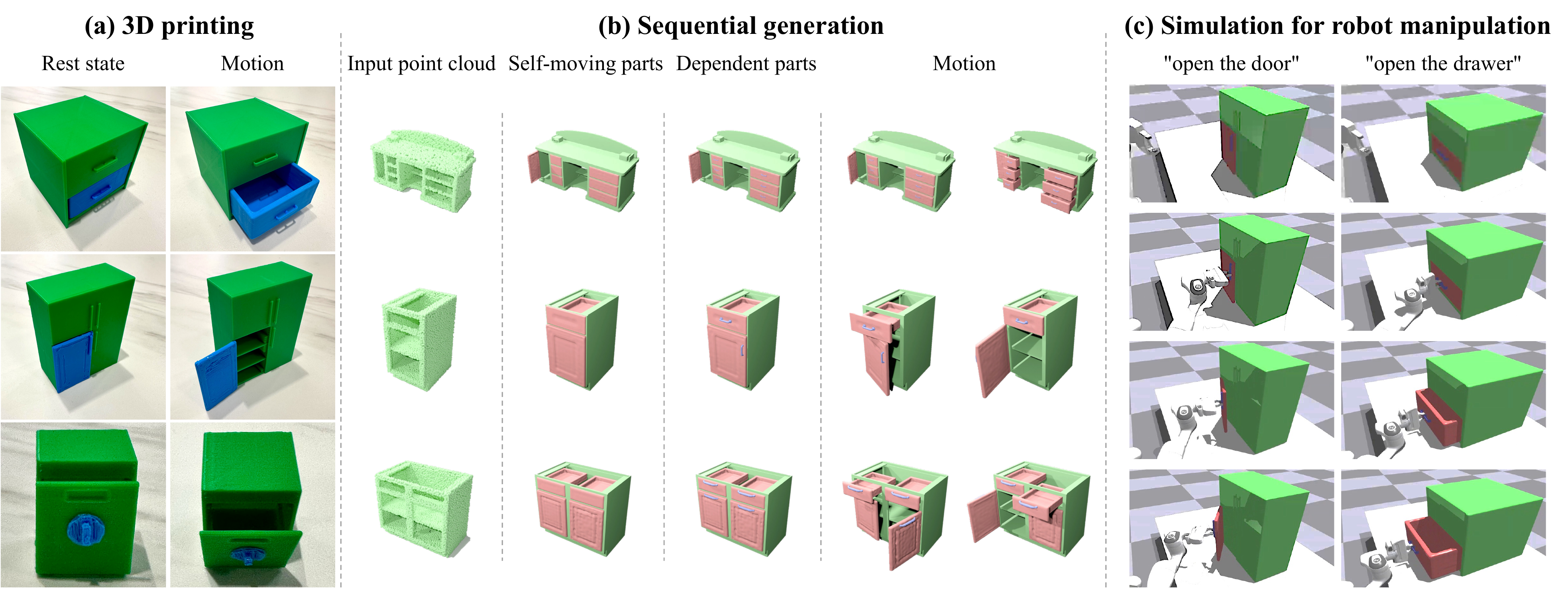}
    \centering
    \vspace{-1mm}
    \caption{\textbf{Downstream applications of our method:} (a) 3D printing for real-world object completion, (b) sequential part generation for complex structures, and (c) simulation for robot manipulation.}
    \vspace{-6mm}
    \label{fig:application}
\end{figure*}

\subsection{Physics-Plausible Generation}

\myparagraph{Metrics.} We introduce a \textit{physical plausibility} metric to evaluate the viability of generated shapes. As illustrated in Algorithm~\ref{algo:physical-plausible}, we first preprocess the meshes, excluding those with multiple large connected parts or those that are not watertight as physically infeasible. For the remaining meshes, a grid search around the part’s ground truth position is conducted to locate positions where both collision loss (\cref{eq:collision-loss}) and contact loss (\cref{eq:contact-loss}) fall below a specified threshold. A part is deemed physically plausible if its movement along the $+x$ and $-x$ axes (for slider drawers) or along the $-z$ axis (for other part categories) causes these losses to exceed the threshold. Notably, for hinge doors, a shared rotation axis is used during loss-guided sampling, while the ground truth rotation axis is used for evaluation. Besides, we also employ the Chamfer Distance (CD) and F-score~\cite{F-score} (@2\%) for evaluation.

\myparagraph{Self-moving Part Generation.}
As shown in \cref{tab:quantative}, our method significantly outperforms the baseline in terms of both generation quality and physical plausibility on generating self-moving parts. The results also highlight the essential role of our proposed physical guidance in the generation process. Notably, as illustrated in \cref{fig:drawer-door}, while the baseline method or the absence of physical guidance often leads to collisions or misalignments at the interface between the object and the generated part, our approach consistently delivers parts that seamlessly integrate with the original object, establishing a robust foundation for subsequent dependent part generation.

\myparagraph{Dependent Part Generation.} As shown in \cref{tab:quantative}, our approach continues to markedly surpass the baseline on dependent parts. \cref{fig:handle-knob} reveals that while the baseline method frequently produces a degenerated distribution (such as generating oval-shaped handles regardless of input or producing broken knobs), our method achieves consistent high-fidelity results that meet the physical requirements.

\begin{figure}[!t]
    \centering
    \vspace{-1.5mm}
    \includegraphics[width=\linewidth]{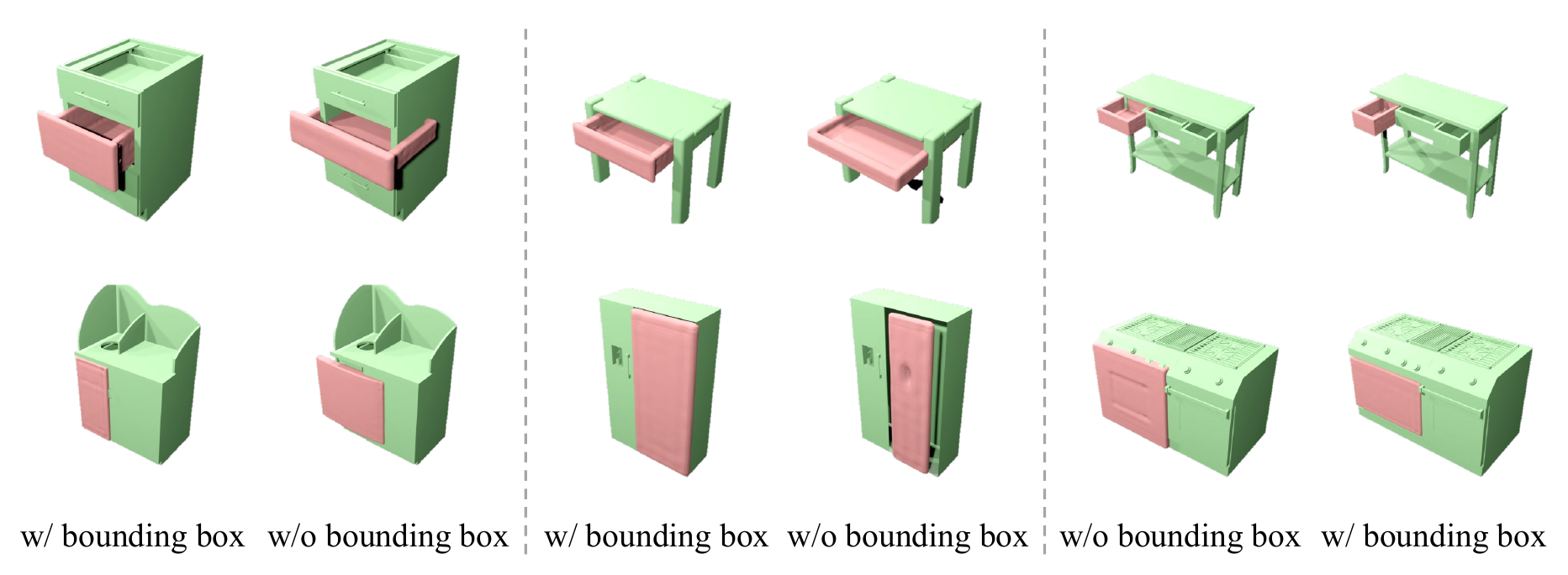}
    \vspace{-6mm}
    \caption{\textbf{Comparison of incorporating the bounding box as condition.} Adding the bounding box alongside the point cloud improves the alignment of the generated part with the target object.}
    \vspace{1mm}
    \label{fig:ablation-bbox}
\end{figure}

\subsection{Downstream Applications}

\myparagraph{3D Printing.}
To better illustrate the effectiveness and physical plausibility of our model, we 3D-printed the generated meshes. The printed parts not only fit seamlessly into the corresponding articulated objects but also demonstrated highly effective motions. Figure~\ref{fig:application}(a) shows these 3D-printed shapes, which were scaled down to $\frac{1}{200}$ of their original size, resulting in dimensions of approximately $10 \text{cm} \times 10 \text{cm} \times 10 \text{cm}$.


\myparagraph{Sequential Part Generation.} We then explore our method's potential in sequential part generation. For an object with multiple missing parts, we first generate the self-moving parts using the ground truth bounding box of each missing part as the condition simultaneously. Once all self-moving parts are generated, they are assembled with the original object. A point cloud is then re-sampled from this assembly and serves as the condition to generate the dependent parts.

\cref{fig:application}(b) shows our results. Although only trained on objects with one missing part, our model presents outstanding generalizability to objects with multiple missing parts. This capability allows for the sequential part generation that adheres to physical constraints, showcasing real-world applications such as obtaining a hierarchy of physically plausible parts given only bounding box-level design.

\myparagraph{Simulation for Robot Manipulation.} 
To further validate the physical plausibility and quality of our generated object, we demonstrate that the generated parts can be manipulated using simple heuristics. Following the setup in \cite{geng2022gapartnet, geng2024sage}, we load the objects in URDF format into the IsaacGym simulator~\cite{makoviychuk2021isaac} and apply a heuristic policy based on the estimated pose. \cref{fig:application}(c) shows that our generated shapes exhibit high quality and can be successfully manipulated.

\subsection{Ablation Studies and Analysis}
This section delves into the effectiveness of different model components through ablation studies. 

\myparagraph{Bounding Box Condition.} We first verify the design of adding the bounding box condition to the diffusion model. As shown by the results in \cref{tab:ablation} and \cref{fig:ablation-bbox}, 
relying solely on the point cloud condition often yields part generation results that fail to fit the input object, such as drawers colliding or doors not filling the whole empty space. Conversely, incorporating the bounding box as an auxiliary condition significantly improves the overall performance of our model.

\myparagraph{Pose Proposal Model.} We then verify the effectiveness of adopting the diffusion-based pose proposal model. As also shown in \cref{tab:ablation}, leveraging the diffusion-based structure produces object bounding boxes of higher accuracy, thus benefiting the overall part generation quality.



\section{Conclusions} 
\label{sec:conclusions}

In this paper, we propose a diffusion-based framework that integrates geometric conditioning and physical constraints, ensuring high accuracy and fidelity of the generated parts in shape and function under real-world interactions. Our results demonstrate substantial improvements in shape quality and physical plausibility over existing methods. These enhancements significantly benefit downstream applications such as 3D printing and robot manipulation, where precision is crucial for the tasks' success.


\bibliographystyle{splncs04}
\bibliography{ref}

\begin{thebibliography}{10}
\providecommand{\url}[1]{\texttt{#1}}
\providecommand{\urlprefix}{URL }
\providecommand{\doi}[1]{https://doi.org/#1}

\bibitem{achlioptas2018learning}
Achlioptas, P., Diamanti, O., Mitliagkas, I., Guibas, L.: Learning representations and generative models for 3d point clouds. In: International Conference on Machine Learning (ICML) (2018)

\bibitem{chen2023urdformer}
Chen, Q., Memmel, M., Fang, A., Walsman, A., Fox, D., Gupta, A.: Urdformer: Constructing interactive realistic scenes from real images via simulation and generative modeling. In: Towards Generalist Robots: Learning Paradigms for Scalable Skill Acquisition @ CoRL 2023 (2023)

\bibitem{chen2019unpaired}
Chen, X., Chen, B., Mitra, N.J.: Unpaired point cloud completion on real scans using adversarial training. In: International Conference on Learning Representations (ICLR) (2020)

\bibitem{cheng2023learning}
Cheng, K., Wu, R., Shen, Y., Ning, C., Zhan, G., Dong, H.: Learning environment-aware affordance for 3d articulated object manipulation under occlusions. In: Advances in Neural Information Processing Systems (NeurIPS) (2023)

\bibitem{cheng2023sdfusion}
Cheng, Y.C., Lee, H.Y., Tulyakov, S., Schwing, A.G., Gui, L.Y.: Sdfusion: Multimodal 3d shape completion, reconstruction, and generation. In: Conference on Computer Vision and Pattern Recognition (CVPR) (2023)

\bibitem{chibane2020implicit}
Chibane, J., Alldieck, T., Pons-Moll, G.: Implicit functions in feature space for 3d shape reconstruction and completion. In: Conference on Computer Vision and Pattern Recognition (CVPR) (2020)

\bibitem{dai2020sg}
Dai, A., Diller, C., Nie{\ss}ner, M.: Sg-nn: Sparse generative neural networks for self-supervised scene completion of rgb-d scans. In: Conference on Computer Vision and Pattern Recognition (CVPR) (2020)

\bibitem{dai2019scan2mesh}
Dai, A., Nie{\ss}ner, M.: Scan2mesh: From unstructured range scans to 3d meshes. In: Conference on Computer Vision and Pattern Recognition (CVPR) (2019)

\bibitem{dai2017shape}
Dai, A., Ruizhongtai~Qi, C., Nie{\ss}ner, M.: Shape completion using 3d-encoder-predictor cnns and shape synthesis. In: Conference on Computer Vision and Pattern Recognition (CVPR) (2017)

\bibitem{deng2023nerdi}
Deng, C., Jiang, C., Qi, C.R., Yan, X., Zhou, Y., Guibas, L., Anguelov, D., et~al.: Nerdi: Single-view nerf synthesis with language-guided diffusion as general image priors. In: Conference on Computer Vision and Pattern Recognition (CVPR) (2023)

\bibitem{deng2024banana}
Deng, C., Lei, J., Shen, W.B., Daniilidis, K., Guibas, L.J.: Banana: Banach fixed-point network for pointcloud segmentation with inter-part equivariance. In: Advances in Neural Information Processing Systems (NeurIPS) (2024)

\bibitem{ding2024opendor}
Ding, Y., Geng, H., Xu, C., Fang, X., Zhang, J., Dai, Q., Wei, S., Zhang, Z., Wang, H.: Open6{DOR}: Benchmarking open-instruction 6-dof object rearrangement and a {VLM}-based approach. In: Robotics: Science and Systems (RSS) (2024)

\bibitem{firman2016structured}
Firman, M., Mac~Aodha, O., Julier, S., Brostow, G.J.: Structured prediction of unobserved voxels from a single depth image. In: Conference on Computer Vision and Pattern Recognition (CVPR) (2016)

\bibitem{geng2023partmanip}
Geng, H., Li, Z., Geng, Y., Chen, J., Dong, H., Wang, H.: Partmanip: Learning cross-category generalizable part manipulation policy from point cloud observations. In: Conference on Computer Vision and Pattern Recognition (CVPR) (2023)

\bibitem{geng2024sage}
Geng, H., Wei, S., Deng, C., Shen, B., Wang, H., Guibas, L.: Sage: Bridging semantic and actionable parts for generalizable articulated-object manipulation under language instructions. In: Robotics: Science and Systems (RSS) (2024)

\bibitem{geng2022gapartnet}
Geng, H., Xu, H., Zhao, C., Xu, C., Yi, L., Huang, S., Wang, H.: Gapartnet: Cross-category domain-generalizable object perception and manipulation via generalizable and actionable parts. In: Conference on Computer Vision and Pattern Recognition (CVPR) (2023)

\bibitem{geng2022end}
Geng, Y., An, B., Geng, H., Chen, Y., Yang, Y., Dong, H.: End-to-end affordance learning for robotic manipulation. In: International Conference on Robotics and Automation (ICRA) (2023)

\bibitem{gong2023arnold}
Gong, R., Huang, J., Zhao, Y., Geng, H., Gao, X., Wu, Q., Ai, W., Zhou, Z., Terzopoulos, D., Zhu, S.C., et~al.: Arnold: A benchmark for language-grounded task learning with continuous states in realistic 3d scenes. In: International Conference on Computer Vision (ICCV) (2023)

\bibitem{han2017high}
Han, X., Li, Z., Huang, H., Kalogerakis, E., Yu, Y.: High-resolution shape completion using deep neural networks for global structure and local geometry inference. In: International Conference on Computer Vision (ICCV) (2017)

\bibitem{ho2022classifier}
Ho, J., Salimans, T.: Classifier-free diffusion guidance. arXiv preprint arXiv:2207.12598  (2022)

\bibitem{hu2024topology}
Hu, J., Fei, B., Xu, B., Hou, F., Yang, W., Wang, S., Lei, N., Qian, C., He, Y.: Topology-aware latent diffusion for 3d shape generation. arXiv preprint arXiv:2401.17603  (2024)

\bibitem{huang2024enhancing}
Huang, Q.: Enhancing implicit shape generators using topological regularizations. In: Joint Mathematics Meetings (2024)

\bibitem{huang2023diffusion}
Huang, S., Wang, Z., Li, P., Jia, B., Liu, T., Zhu, Y., Liang, W., Zhu, S.C.: Diffusion-based generation, optimization, and planning in 3d scenes. In: Conference on Computer Vision and Pattern Recognition (CVPR) (2023)

\bibitem{jiang2022ditto}
Jiang, Z., Hsu, C.C., Zhu, Y.: Ditto: Building digital twins of articulated objects from interaction. In: Conference on Computer Vision and Pattern Recognition (CVPR) (2022)

\bibitem{kazhdan2006poisson}
Kazhdan, M., Bolitho, M., Hoppe, H.: Poisson surface reconstruction. In: Proceedings of Eurographics Symposium on Geometry Processing (2006)

\bibitem{kazhdan2013screened}
Kazhdan, M., Hoppe, H.: Screened poisson surface reconstruction. ACM Transactions on Graphics (TOG)  \textbf{32}(3),  1--13 (2013)

\bibitem{kim2012acquiring}
Kim, Y.M., Mitra, N.J., Yan, D.M., Guibas, L.: Acquiring 3d indoor environments with variability and repetition. ACM Transactions on Graphics (TOG)  \textbf{31}(6),  1--11 (2012)

\bibitem{lei2023nap}
Lei, J., Deng, C., Shen, B., Guibas, L., Daniilidis, K.: Nap: Neural 3d articulation prior. arXiv preprint arXiv:2305.16315  (2023)

\bibitem{li2024ag2maniplearningnovelmanipulation}
Li, P., Liu, T., Li, Y., Han, M., Geng, H., Wang, S., Zhu, Y., Zhu, S.C., Huang, S.: Ag2manip: Learning novel manipulation skills with agent-agnostic visual and action representations. arXiv preprint arXiv:2404.17521  (2024)

\bibitem{li2020category}
Li, X., Wang, H., Yi, L., Guibas, L.J., Abbott, A.L., Song, S.: Category-level articulated object pose estimation. In: Conference on Computer Vision and Pattern Recognition (CVPR) (2020)

\bibitem{li2015database}
Li, Y., Dai, A., Guibas, L., Nie{\ss}ner, M.: Database-assisted object retrieval for real-time 3d reconstruction. In: Computer Graphics Forum (2015)

\bibitem{lin2023magic3d}
Lin, C.H., Gao, J., Tang, L., Takikawa, T., Zeng, X., Huang, X., Kreis, K., Fidler, S., Liu, M.Y., Lin, T.Y.: Magic3d: High-resolution text-to-3d content creation. In: Conference on Computer Vision and Pattern Recognition (CVPR) (2023)

\bibitem{liu2023semi}
Liu, G., Sun, Q., Huang, H., Ma, C., Guo, Y., Yi, L., Huang, H., Hu, R.: Semi-weakly supervised object kinematic motion prediction. In: Conference on Computer Vision and Pattern Recognition (CVPR) (2023)

\bibitem{liu2024survey}
Liu, J., Savva, M., Mahdavi-Amiri, A.: Survey on modeling of articulated objects. arXiv preprint arXiv:2403.14937  (2024)

\bibitem{liu2024cage}
Liu, J., Tam, H.I.I., Mahdavi-Amiri, A., Savva, M.: Cage: Controllable articulation generation. In: Conference on Computer Vision and Pattern Recognition (CVPR) (2024)

\bibitem{liu2024one}
Liu, M., Xu, C., Jin, H., Chen, L., Varma~T, M., Xu, Z., Su, H.: One-2-3-45: Any single image to 3d mesh in 45 seconds without per-shape optimization. In: Advances in Neural Information Processing Systems (NeurIPS) (2023)

\bibitem{liu2023zero}
Liu, R., Wu, R., Van~Hoorick, B., Tokmakov, P., Zakharov, S., Vondrick, C.: Zero-1-to-3: Zero-shot one image to 3d object. In: International Conference on Computer Vision (ICCV) (2023)

\bibitem{makoviychuk2021isaac}
Makoviychuk, V., Wawrzyniak, L., Guo, Y., Lu, M., Storey, K., Macklin, M., Hoeller, D., Rudin, N., Allshire, A., Handa, A., State, G.: Isaac gym: High performance gpu-based physics simulation for robot learning. arXiv preprint arXiv:2108.10470  (2021)

\bibitem{mezghanni2022physical}
Mezghanni, M., Bodrito, T., Boulkenafed, M., Ovsjanikov, M.: Physical simulation layer for accurate 3d modeling. In: Conference on Computer Vision and Pattern Recognition (CVPR) (2022)

\bibitem{mezghanni2021physically}
Mezghanni, M., Boulkenafed, M., Lieutier, A., Ovsjanikov, M.: Physically-aware generative network for 3d shape modeling. In: Conference on Computer Vision and Pattern Recognition (CVPR) (2021)

\bibitem{mitra2006partial}
Mitra, N.J., Guibas, L.J., Pauly, M.: Partial and approximate symmetry detection for 3d geometry. ACM Transactions on Graphics (TOG)  \textbf{25}(3),  560--568 (2006)

\bibitem{mittal2022autosdf}
Mittal, P., Cheng, Y.C., Singh, M., Tulsiani, S.: Autosdf: Shape priors for 3d completion, reconstruction and generation. In: Conference on Computer Vision and Pattern Recognition (CVPR) (2022)

\bibitem{mu2021sdf}
Mu, J., Qiu, W., Kortylewski, A., Yuille, A., Vasconcelos, N., Wang, X.: A-sdf: Learning disentangled signed distance functions for articulated shape representation. In: International Conference on Computer Vision (ICCV) (2021)

\bibitem{nan2012search}
Nan, L., Xie, K., Sharf, A.: A search-classify approach for cluttered indoor scene understanding. ACM Transactions on Graphics (TOG)  \textbf{31}(6),  1--10 (2012)

\bibitem{nealen2006laplacian}
Nealen, A., Igarashi, T., Sorkine, O., Alexa, M.: Laplacian mesh optimization. In: Proceedings of International Conference on Computer Graphics and Interactive Techniques in Australasia and Southeast Asia (2006)

\bibitem{nguyen2016field}
Nguyen, D.T., Hua, B.S., Tran, K., Pham, Q.H., Yeung, S.K.: A field model for repairing 3d shapes. In: Conference on Computer Vision and Pattern Recognition (CVPR) (2016)

\bibitem{pauly2008discovering}
Pauly, M., Mitra, N.J., Wallner, J., Pottmann, H., Guibas, L.J.: Discovering structural regularity in 3d geometry. ACM Transactions on Graphics (TOG)  \textbf{27}(3),  1--11 (2008)

\bibitem{poole2022dreamfusion}
Poole, B., Jain, A., Barron, J.T., Mildenhall, B.: Dreamfusion: Text-to-3d using 2d diffusion. arXiv preprint arXiv:2209.14988  (2022)

\bibitem{qi2017pointnet++}
Qi, C.R., Yi, L., Su, H., Guibas, L.J.: Pointnet++: Deep hierarchical feature learning on point sets in a metric space. In: Advances in Neural Information Processing Systems (NeurIPS) (2017)

\bibitem{rao2022patchcomplete}
Rao, Y., Nie, Y., Dai, A.: Patchcomplete: Learning multi-resolution patch priors for 3d shape completion on unseen categories. In: Advances in Neural Information Processing Systems (NeurIPS) (2022)

\bibitem{shi2023zero123++}
Shi, R., Chen, H., Zhang, Z., Liu, M., Xu, C., Wei, X., Chen, L., Zeng, C., Su, H.: Zero123++: a single image to consistent multi-view diffusion base model. arXiv preprint arXiv:2310.15110  (2023)

\bibitem{sipiran2014approximate}
Sipiran, I., Gregor, R., Schreck, T.: Approximate symmetry detection in partial 3d meshes. In: Computer Graphics Forum (2014)

\bibitem{smith2017improved}
Smith, E.J., Meger, D.: Improved adversarial systems for 3d object generation and reconstruction. In: Conference on Robot Learning (CoRL) (2017)

\bibitem{cvae}
Sohn, K., Lee, H., Yan, X.: Learning structured output representation using deep conditional generative models. In: Advances in Neural Information Processing Systems (NeurIPS) (2015)

\bibitem{song2020denoising}
Song, J., Meng, C., Ermon, S.: Denoising diffusion implicit models. In: International Conference on Learning Representations (ICLR) (2021)

\bibitem{song2017semantic}
Song, S., Yu, F., Zeng, A., Chang, A.X., Savva, M., Funkhouser, T.: Semantic scene completion from a single depth image. In: Conference on Computer Vision and Pattern Recognition (CVPR) (2017)

\bibitem{sorkine2004least}
Sorkine, O., Cohen-Or, D.: Least-squares meshes. In: Proceedings Shape Modeling Applications (2004)

\bibitem{speciale2016symmetry}
Speciale, P., Oswald, M.R., Cohen, A., Pollefeys, M.: A symmetry prior for convex variational 3d reconstruction. In: European Conference on Computer Vision (ECCV) (2016)

\bibitem{sung2015data}
Sung, M., Kim, V.G., Angst, R., Guibas, L.: Data-driven structural priors for shape completion. ACM Transactions on Graphics (TOG)  \textbf{34}(6),  1--11 (2015)

\bibitem{F-score}
Tatarchenko, M., Richter, S.R., Ranftl, R., Li, Z., Koltun, V., Brox, T.: What do single-view 3d reconstruction networks learn? In: Conference on Computer Vision and Pattern Recognition (CVPR) (2019)

\bibitem{thrun2005shape}
Thrun, S., Wegbreit, B.: Shape from symmetry. In: International Conference on Computer Vision (ICCV) (2005)

\bibitem{tseng2022cla}
Tseng, W.C., Liao, H.J., Yen-Chen, L., Sun, M.: Cla-nerf: Category-level articulated neural radiance field. In: International Conference on Robotics and Automation (ICRA) (2022)

\bibitem{wang2021synthesizing}
Wang, J., Xu, H., Xu, J., Liu, S., Wang, X.: Synthesizing long-term 3d human motion and interaction in 3d scenes. In: Conference on Computer Vision and Pattern Recognition (CVPR) (2021)

\bibitem{wu2020multimodal}
Wu, R., Chen, X., Zhuang, Y., Chen, B.: Multimodal shape completion via conditional generative adversarial networks. In: European Conference on Computer Vision (ECCV) (2020)

\bibitem{xu2023neurallift}
Xu, D., Jiang, Y., Wang, P., Fan, Z., Wang, Y., Wang, Z.: Neurallift-360: Lifting an in-the-wild 2d photo to a 3d object with 360deg views. In: Conference on Computer Vision and Pattern Recognition (CVPR) (2023)

\bibitem{xu2022universal}
Xu, Z., He, Z., Song, S.: Universal manipulation policy network for articulated objects. IEEE Robotics and Automation Letters (RA-L)  \textbf{7}(2),  2447--2454 (2022)

\bibitem{yang2024physcene}
Yang, Y., Jia, B., Zhi, P., Huang, S.: Physcene: Physically interactable 3d scene synthesis for embodied ai. In: Conference on Computer Vision and Pattern Recognition (CVPR) (2024)

\bibitem{yi2018deep}
Yi, L., Huang, H., Liu, D., Kalogerakis, E., Su, H., Guibas, L.: Deep part induction from articulated object pairs. arXiv preprint arXiv:1809.07417  (2018)

\bibitem{yu2024dexgraspnet}
Yu, X., Zhang, J., Liu, H., Li, D., Geng, H., Wang, H., Ding, Y., Chen, J.: Dexgraspnet 2.0: Learning generative dexterous grasping in large-scale synthetic cluttered scenes. In: 2nd Workshop on Dexterous Manipulation: Design, Perception and Control (RSS) (2024)

\bibitem{yu2021pointr}
Yu, X., Rao, Y., Wang, Z., Liu, Z., Lu, J., Zhou, J.: Pointr: Diverse point cloud completion with geometry-aware transformers. In: International Conference on Computer Vision (ICCV) (2021)

\bibitem{yuan2023physdiff}
Yuan, Y., Song, J., Iqbal, U., Vahdat, A., Kautz, J.: Physdiff: Physics-guided human motion diffusion model. In: International Conference on Computer Vision (ICCV) (2023)

\bibitem{zhang2023genpose}
Zhang, J., Wu, M., Dong, H.: Genpose: Generative category-level object pose estimation via diffusion models. In: Advances in Neural Information Processing Systems (NeurIPS) (2023)

\bibitem{zhang2021unsupervised}
Zhang, J., Chen, X., Cai, Z., Pan, L., Zhao, H., Yi, S., Yeo, C.K., Dai, B., Loy, C.C.: Unsupervised 3d shape completion through gan inversion. In: Conference on Computer Vision and Pattern Recognition (CVPR) (2021)

\bibitem{zhang2020place}
Zhang, S., Zhang, Y., Ma, Q., Black, M.J., Tang, S.: Place: Proximity learning of articulation and contact in 3d environments. In: International Conference on 3D Vision (3DV) (2020)

\bibitem{zhang2020generating}
Zhang, Y., Hassan, M., Neumann, H., Black, M.J., Tang, S.: Generating 3d people in scenes without people. In: Conference on Computer Vision and Pattern Recognition (CVPR) (2020)

\bibitem{zhao2007robust}
Zhao, W., Gao, S., Lin, H.: A robust hole-filling algorithm for triangular mesh. The Visual Computer  \textbf{23},  987--997 (2007)

\bibitem{zheng2022sdf}
Zheng, X., Liu, Y., Wang, P., Tong, X.: Sdf-stylegan: Implicit sdf-based stylegan for 3d shape generation. In: Computer Graphics Forum (2022)

\bibitem{zhou2023sparsefusion}
Zhou, Z., Tulsiani, S.: Sparsefusion: Distilling view-conditioned diffusion for 3d reconstruction. In: Conference on Computer Vision and Pattern Recognition (CVPR) (2023)

\end{thebibliography}

\end{document}